\documentclass[nohyperref]{article}

\usepackage{microtype}
\usepackage{graphicx}
\usepackage{subfigure}
\usepackage{booktabs} % for professional tables
\usepackage{multirow}
\usepackage{hyperref}

\usepackage[accepted]{icml2022}

\usepackage{amsmath}
\usepackage{amssymb}
\usepackage{mathtools}
\usepackage{amsthm}
\usepackage[capitalize,noabbrev]{cleveref}

\theoremstyle{plain}

\theoremstyle{definition}

\theoremstyle{remark}

\usepackage[textsize=tiny]{todonotes}

\icmltitlerunning{Standard Deviation-Based Quantization for DNNs}

\begin{document}

\twocolumn[
\icmltitle{Standard Deviation-Based Quantization for Deep Neural Networks}



\icmlsetsymbol{equal}{*}

\begin{icmlauthorlist}
\icmlauthor{Amir Ardakani}{}
\icmlauthor{Arash Ardakani}{}
\icmlauthor{Brett Meyer}{}
\icmlauthor{James J. Clark}{}
\icmlauthor{Warren J. Gross}{} \\ 
\icmlauthor{ }{} \\
\icmlauthor{Department of Electrical and Computer Engineering, McGill University, Montreal, QC, Canada}{}

%\icmlauthor{Firstname6 Lastname6}{sch,yyy,comp}
%\icmlauthor{Firstname7 Lastname7}{comp}
%\icmlauthor{Firstname8 Lastname8}{sch}
%\icmlauthor{Firstname8 Lastname8}{yyy,comp}
\end{icmlauthorlist}

%\icmlaffiliation{yyy}{Department of Electrical and Computer Engineering, McGill University, Montreal, QC, Canada}
%\icmlaffiliation{comp}{Company Name, Location, Country}
%\icmlaffiliation{sch}{School of ZZZ, Institute of WWW, Location, Country}

\icmlcorrespondingauthor{Amir Ardakani}{amir.ardakani@mail.mcgill.ca}
%\icmlcorrespondingauthor{Firstname2 Lastname2}{first2.last2@www.uk}

\icmlkeywords{Quantization, Deep neural networks, Standard deviation, Pruning}

\vskip 0.3in
]




\begin{abstract}
Quantization of deep neural networks is a promising approach that reduces the inference cost, making it feasible to run deep networks on resource-restricted devices. Inspired by existing methods, we propose a new framework to learn the quantization intervals (discrete values) using the knowledge of the network's weight and activation distributions, i.e., standard deviation. Furthermore, we propose a novel base-2 logarithmic quantization scheme to quantize weights to power-of-two discrete values. Our proposed scheme allows us to replace resource-hungry high-precision multipliers with simple shift-add operations. According to our evaluations, our method outperforms existing work on CIFAR10 and ImageNet datasets and even achieves better accuracy performance with 3-bit weights and activations when compared to the full-precision models. Moreover, our scheme simultaneously prunes the network's parameters and allows us to flexibly adjust the pruning ratio during the quantization process.
\end{abstract}

%\vspace{-0.8cm}
\section{Introduction}
\label{intro}
Deep neural networks (DNNs) have shown promising results in various applications including Image classification and language processing tasks. However, deployment of the current state-of-the-art (SOTA) DNNs requires expensive and powerful hardware (e.g., GPUs) to perform costly high-precision multiplications \cite{binaryconnect}. Moreover, DNNs are generally over-parameterized and utilize large memories, thereby require heavy data transmission between the computation and memory units \cite{shiftaddnet}. When it comes to resource-constrained hardware such as mobile devices and embedded systems, the aforementioned requirements make the deployment of such networks a challenging task. Correspondingly, a large amount of effort has gone into developing algorithms and hardware, aiming to reduce the deployment costs of DNNs, whilst maintaining high accuracy. Efficient architecture design \cite{squeezenext, mobilenetv2, mobilenetv3}, pruning \cite{prunquant, prun2}, stochastic computing \cite{ardakani2017vlsi, gross2019stochastic} and spiking neural networks \cite{ghosh2009spiking, smithson2016stochastic} are examples of such efforts.    

Another realm of studies focuses on quantization of DNNs where real values are mapped to discrete integer values \cite{dorefa}. Using integer values at inference dramatically reduces the computational and memory costs. In this work, we also focus on the quantization of DNNs. Although most existing works have employed uniform quantization methods where discrete levels have the same step size \cite{pact, lsq, qil}, DNNs can be quantized to non-uniform discrete integers as well \cite{lqnet, miyashita2016convolutional, s3, zhou2017incremental}.\par
%To quantize DNNs, we can either map the real values to the discrete levels that \textbf{accurately represents} the data or we can find the quantization intervals that produce the \textbf{best} metrics. Most recent work based on the former quantization approach often aim to achieve accurate representation of data by minimizing the quantization error between the quantizer's input and output. On the other hand, methods based on the latter approach try to learn the quantization intervals (i.e., discrete levels) that are best to reduce the task loss. Minimizing the quantization error, as accurately as possible, does not necessary achieve the best metrics. In fact, the existing empirical results have shown that methods based on learning the quantization intervals are superior than the ones minimizing the quantization errors. Building on this observation, we propose a new quantization framework capable of learning the quantization intervals while optimizing the task loss. %In this work, we make the following contributions:
In general, the quantization methods mainly fall into three categories. The first category includes the methods trying to minimize the quantization error where real values are mapped to the discrete levels that accurately represents the data \cite{ternary, HWGQ, tsq, lqnet, llsq}. Most recent works that fall into the first category often aim to achieve accurate representation of data by minimizing the quantization error between the quantizer's input and output. These methods require to constantly monitor weights and activations values (online) or to use the structural characterization of data from the full-precision network (offline) for minimizing the quantization error. For instance, in the HWGQ scheme \cite{HWGQ}, the quantization error is minimized using Lloyd’s algorithm \cite{lloyd1982least} and the statistical structure (distribution) of the activations obtained from the full-precision network to fit the quantizer to the data. The main drawback of this method is its offline optimization and that the distribution of the full-precision activations is not necessarily the same as the quantized network. Contrary to HWGQ, LLSQ method uses an online algorithm to minimize the quantization error during the training phase \cite{llsq}. LQ-Net is another online method \cite{lqnet}, where a non-uniform quantization scheme was developed that transforms the quantization error to a linear regression problem with a closed form solution. The performance of HWGQ, LLSQ and LQ-Net has been outperformed by other methods (i.e., the second category), empirically showing that minimizing the quantization error is not the best approach to obtain quantized networks. \par
In the second category, the quantization is treated as an optimization problem that can be simultaneously optimized with the loss function during the training phase. More specifically, the quantizers are redesigned with learnable parameters that can be then used to find the optimal quantization
intervals. In this approach, the goal is to find the optimizer that can produce the best accuracy performance, whereas in the first category, the goal is to find the optimizer that represents the data more accurately (less quantization error). PACT \cite{pact}, QIL \cite{qil} and LSQ \cite{lsq} are the well-known quantization methods from this category.  \par

Finally, the third category mainly focuses on the training process of the quantized networks \cite{zhou2017incremental, PQTSG}. This includes techniques and tricks that are used during the training of the quantized network to improve its performance accuracy. For instance, it has been shown that a progressive training framework with the help of knowledge distillation significantly improves the performance of the DoReFa-Net \cite{dorefa} quantization method on various tasks \cite{PQTSG}. \par

In this work, we aim to close the gap between these methods by exploiting the statistical characterization of weights and activations distribution during the quantization process. In other words, we propose a new quantization scheme that utilizes the standard deviation of weights/activations distribution during the training phase to find the optimal optimizer that produces the best accuracy performance using task loss and back-propagation. We also explain how standard deviation contributes in improving our quantization method in \cref{contributing}. Furthermore, in \cref{Techniques}, we propose two training techniques which are employed in our proposed quantization method to further improve the performance of the quantized networks. 

Moreover, unlike the existing methods that have only focused on one type of quantization, that is either uniform or non-uniform quantization, we propose a quantization framework that is capable of both uniform and non-uniform quantization. Our non-uniform quantization method uses power-of-two discrete levels. We show that the proposed non-uniform quantization method outperforms the SOTA results \cite{s3} while having $10\times$ faster convergence. 

Quantizers with the discrete intervals that include zero level, inherently benefit from pruning as values that fall into this level during the quantization are set to 0 and can be pruned. This valuable property makes quantization methods more desirable over other cost reduction methods. However, there are only a few work investigating the relation between pruning and quantization, and their impact on each other \cite{qil, prunquant}. In this work, not only we study the pruning property of quantization in more depth, we also show that our method allows us to flexibly adjust the pruning ratio during the quantization process. For instance, we show that up to 40\% of weights in a 3-bit ResNet-18 model can be pruned at the cost of less than 1\% accuracy loss when compared to its full-precision counterpart.  \par
\vspace{-0.2cm}
\section{Standard Deviation-Based Quantization}

To make the connection between the existing quantization methods, we try to solve the following
question: ``How much of the information is important in a quantized network?''. To answer this
question, we need to look at the statistical characterization of weights and activations more carefully. It has been shown that the output of the convolution/fully-connected layers has a bell-shaped
distribution, especially if followed by a batch-normalization layer \cite{ioffe2015batch}. Besides the activations, the weights of a network regularized with L2-regularization method have a bell-shaped distribution as well. The width of a bell-shaped distribution is defined by its standard deviation. For instance, in a Gaussian distribution, the values within three standard deviation ($\sigma$) from the mean ($\mu$) account for 99.73\% of the entire data. However, not all of these values are as important as
the others in a neural network. Here we propose a standard deviation-based clipping function with a
learnable parameter ($\alpha$) which is capable of determining how much of these values is actually important.
The backbone of our proposed clipping function is the PACT \cite{pact} method. In our proposed function, we integrate the standard deviation of the weights/activations in to the PACT such that
\begin{equation}
\label{clip_general}
\text{y}(x) = \left\{\begin{matrix}
x & |x| < \alpha\sigma \\ 
\text{sign}(x)\cdot\alpha\sigma & |x| \geq \alpha\sigma   
\end{matrix}\right.,
\end{equation}
where $\alpha$ and $\sigma$ are the quantizer's learnable parameter and the standard deviation of the weights/activations, respectively. For the activations, we can integrate the ReLU function with our clipping function, i.e.,
\begin{equation}
\label{clip-activation}
\text{y}(x) = \left\{\begin{matrix}
0 & x \leq 0 \\ 
x & 0 < x < \alpha\sigma \\ 
\alpha\sigma & x \geq \alpha\sigma   
\end{matrix}\right..
\end{equation}
The proposed clipping function simply discards the ``outliers'' that are $\alpha\sigma$ far from the mean of the weights/activations distribution. Note that in our clipping function, the mean value of weights is considered to be zero all the time. This is because in our experiments, the mean value of weights was always close to zero and did not impact the results when it was omitted. The mean of the activations, however, is always zero. Note that in \cref{clip-activation}, we use ReLU function to assign zeros to all negative values. Consequently, we should not calculate $\sigma$ directly from the activations input. Instead, we calculate the $\sigma$ by removing all negative values from the inputs and then we apply a horizontal mirroring on the remaining values. It is evident that after this step, the mean value of activations distribution is always zero. \par
Unlike the standard deviation of the weights distribution, the standard deviation of activations can have a significantly different value whenever a new batch of data is introduced to the network. Therefore, we need to average $\sigma$ for activations distribution during the training phase. Inspired by the batch-normalization technique, we keep track of $\sigma$ by employing the exponential running average with the momentum of 0.001,   
\begin{equation}
%\label{runningstd}
\hat{\sigma}_\text{new} = (1 - \text{momentum}) \times \hat{\sigma} + \text{momentum} \times \sigma_t,
\end{equation}
where $\hat{\sigma}$ is the running average and $\sigma_t$ is the standard deviation of the current batch. It is worth mentioning that the momentum value of 0.001 was empirically obtained from our experiments. \par

The output of the clipping function can be then quantized to $L_P+L_N+1$ discrete levels (integer values) using $b$ bits such that
\begin{equation}
\label{yd}
y_d = clip(\left \lfloor{y \cdot  \frac{L_P}{\alpha\sigma}}\right \rceil, -L_N, L_P), 
\end{equation}
where function $\left \lfloor{a}\right \rceil$ rounds $a$ to its nearest integer value and $y_{d} \in \mathbb{N}$. Variable $L$ is determined by the quantizer bit-width ($b$) and the data type (signed/Unsigned). For unsigned data (activations) $L_N = 0$ and $L_P = 2^b-1$ and for signed data (weights) $L_P = L_N = 2^{b-1}-1$. It is worth mentioning that our method quantizes signed values symmetrically. For instance, quantization of signed values using 2 bits results in three discrete values (i.e., ternary discrete values).\par

\begin{figure}[t]
%\vskip 0.2in
\begin{center}
\centerline{\includegraphics[width=0.9\columnwidth]{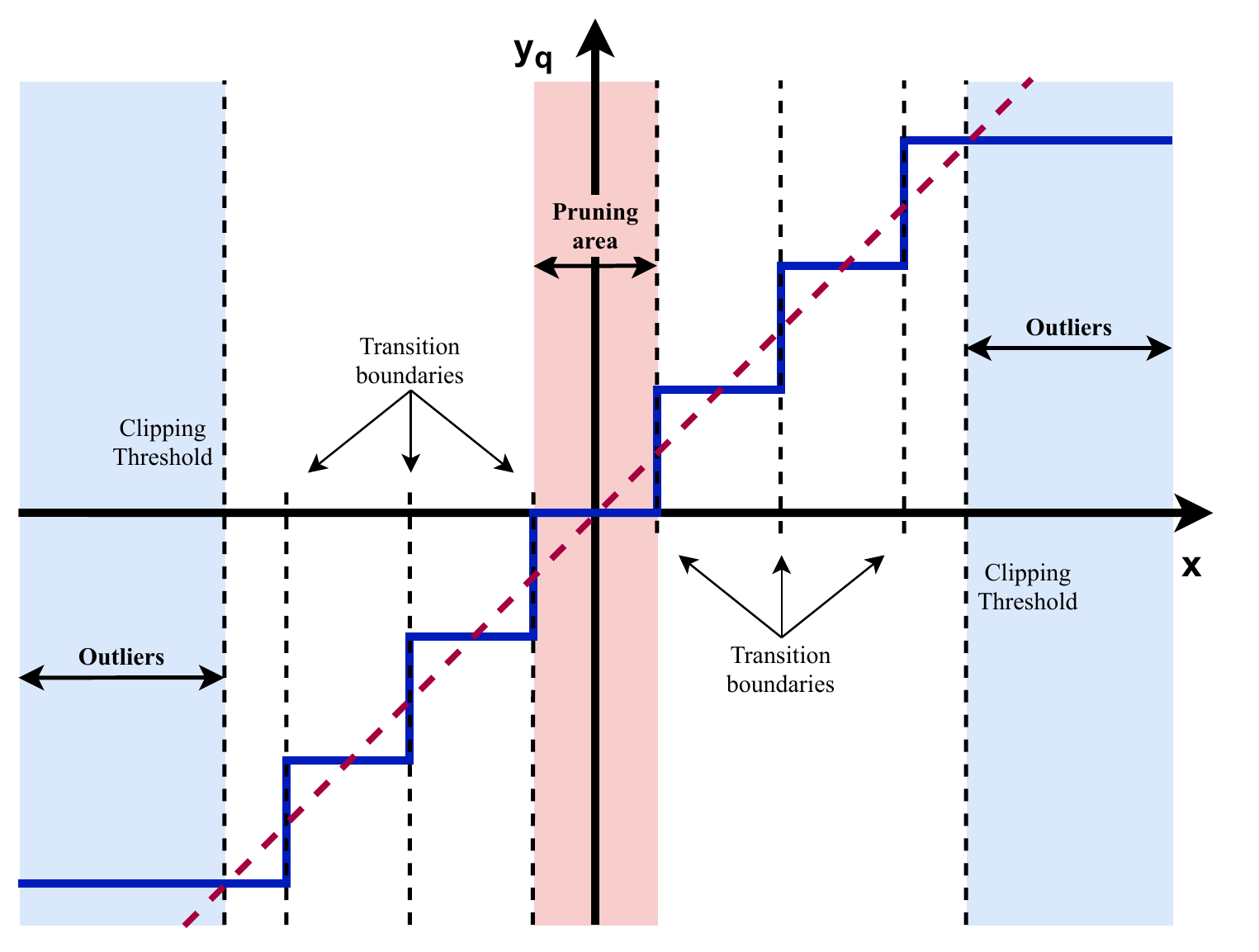}}
\vskip -0.1in
\caption{An example of a 3-bit quantizer for signed values.}
\label{quantization}
\end{center}
\vskip -0.3in
\end{figure}

Finally, the quantization process is completed by multiplying the quantizer scale value with $y_d$ such that
\begin{equation}
\label{yq}
y_q = y_d \cdot \frac{\alpha\sigma}{L_P}, 
\end{equation}
where $y_{q} \in \mathbb{R}$. We would like to emphasize that the multiplication of the quantizer scale with $y_d$ can simply be integrated with the batch normalization layer without any additional costs. 
As shown in \cref{quantization}, our proposed quantization method automatically prunes the network as well. In fact, any values that falls into the \textit{pruning area} of the quantizer is pruned (set to zero) during the quantization. From \cref{yd}, we can observe that 
\begin{equation*}
\label{pruning_area}
\text{if}~~|y| < \frac{\alpha\sigma}{2L_P}\text{, }~~ \text{then}~~ y_d = 0, 
\end{equation*}
which shows a direct relation between the pruning ratio and the clipping threshold of the quantizer. In other words, higher clipping thresholds result in greater pruning ratios. 

\subsection{Optimizing $\alpha$ with Backpropagation}
Following the same principles introduced in PACT and using Straight-Through Estimator (ETS) \cite{bengio2013estimating}, we can derive the gradient for the quantizer parameter $\alpha$ by
\begin{equation}
\label{grad}
g_\alpha = \left\{\begin{matrix}
\sigma \cdot g_y & x \geq \alpha\sigma \\ 
0 & \text{otherwise} \\ 
\end{matrix}\right.,
\end{equation}
where $g_y$ is the incoming gradient from the front layers. Furthermore, the gradient of the activations input $g_x$ can be computed using 
\begin{equation}
\label{lossx}
g_x = \left\{\begin{matrix}
g_y & 0 < x < \alpha\sigma   \\ 
0 & \text{otherwise}    
\end{matrix}\right..
\end{equation}
We can add a gradient scale $s$ and weight decay $\lambda\alpha$ to $g_\alpha$ and rewrite \cref{grad} as
\begin{equation}
\label{newgrad}
g_\alpha = \left\{\begin{matrix}
s \sigma \cdot g_y + \lambda\alpha & x \geq \alpha\sigma \\ 
0 & \text{otherwise} \\ 
\end{matrix}\right..
\end{equation}
The gradient scale and weight decay were added to $g_\alpha$ to prevent $\alpha$ from exploding/vanishing and to control the pruning ratio. 
\vspace{-0.2cm}
\subsection{Base-2 Logarithmic Quantization} \label{shiftnet}
We can use the clipping function described in \cref{clip_general} to quantize weights to power-of-two discrete values, i.e., $y_{p2} \in \{0, \pm2^k\}$, where $k \in \mathbb{Z}^{+}_{0}$. With this new power-of-two quantized weights, we can replace the expensive multiplications with simple shift and addition/subtraction operations. To quantize weights to power-of-two discrete values, we first modify \cref{yd} to produce discrete integers ($y_{int} \in \mathbb{N}$) such that   
\begin{equation}
\label{y_int}
y_{int} = \left \lfloor{\log_2 (|y| \cdot  \frac{L_{p2}}{\alpha\sigma})}\right \rceil, 
\end{equation}
where $L_{p2} = 2^{2^{b-1}-2}$. We then transform $y_{int}$ to discrete power-of-two integers including zero ($y_{p2} \in \{0, \pm2^k\}$) using 
\begin{equation}
\label{y_p2}
y_{p2} = \left\{\begin{matrix}
clip(\text{sign}(y) \cdot 2^{y_{int}}, -L_{p2}, L_{p2})  & y_{int} \geq 0   \\ 
0 & \text{otherwise}    
\end{matrix}\right..
\end{equation}
For instance, using 3 bits ($b = 3$) we can quantize weights to seven discrete values, i.e., $y_{p2}^{3-bit} \in \{0,\pm1, \pm2, \pm4\}$. Similarly, we can use \cref{yq} to multiply the quantizer scale with $y_{p2}$ to complete the quantization process. Finally, we can use the same method described in \cref{newgrad} to optimize the quantizer parameter $\alpha$.
\vspace{-0.2cm}
\section{Contributing Factors} \label{contributing}
The main intuition behind our heuristic approach to parametize the quantizers with statistical structure of the weights/activations is to include a new feature, i.e., the standard deviation of the data distribution during the quantization process. Unlike the previous parameterization methods, the loss optimization process (backpropagation) of weights and the quantizer parameter $\alpha$ of our proposed method, relies on both the input samples and the statistical structure of weights/activations from each layer. For instance, the standard deviation shows how dense one distribution is. In a highly dense distribution (small $\sigma$), where a huge portion of data resides inside the pruning area, clearly we do not want the outliers to move the clipping threshold further from the zero. In this example, the small value of $\sigma$ does not allow the gradients of the outliers to significantly shift the clipping threshold, according the \cref{newgrad}. In addition to the aforementioned explanation, we believe that there are three other contributing factors to why our method works: inclusivity, adaptive gradient scale factor and faster convergence.

\textbf{Inclusivity:} Unlike PACT where the gradients of the quantizer parameter only rely on the outliers, the gradients in our method are sensitive to the entire data. More specifically, we introduce the standard deviation of weights and activations distribution as a new feature that appears in the task loss and is used to scale the quantizer parameter.

\textbf{Adaptive gradient scale factor:} The impact of the gradient scale has been investigated in LSQ. It was shown that large gradient scales (e.g., 1) produce weak accuracy results. Looking at \cref{newgrad}, the standard deviation in our method can also be interpreted as an adaptive scaling factor for the quantizer parameter $\alpha$. In most of our experiments, the gradient scale was initially set to 1 and then adjusted to a proper value that achieves the best accuracy performance. In fact, the role of standard deviation as the adaptive gradient scale enables us to trade-off between the accuracy performance and the pruning ratio by treating the gradient scale as a hyper-parameter. Using a small gradient scale forces $\alpha$ to converge to smaller values and consequently, less data is pruned. On the other hand, setting the gradient scale to larger values allows us to prune more data, though at the cost of loosing accuracy performance. We investigate this property in \cref{Pruningratio}.

\begin{figure}[t]
%\vskip 0.2in
\begin{center}
\centerline{\includegraphics[width=0.7\columnwidth, angle=-90, origin=c]{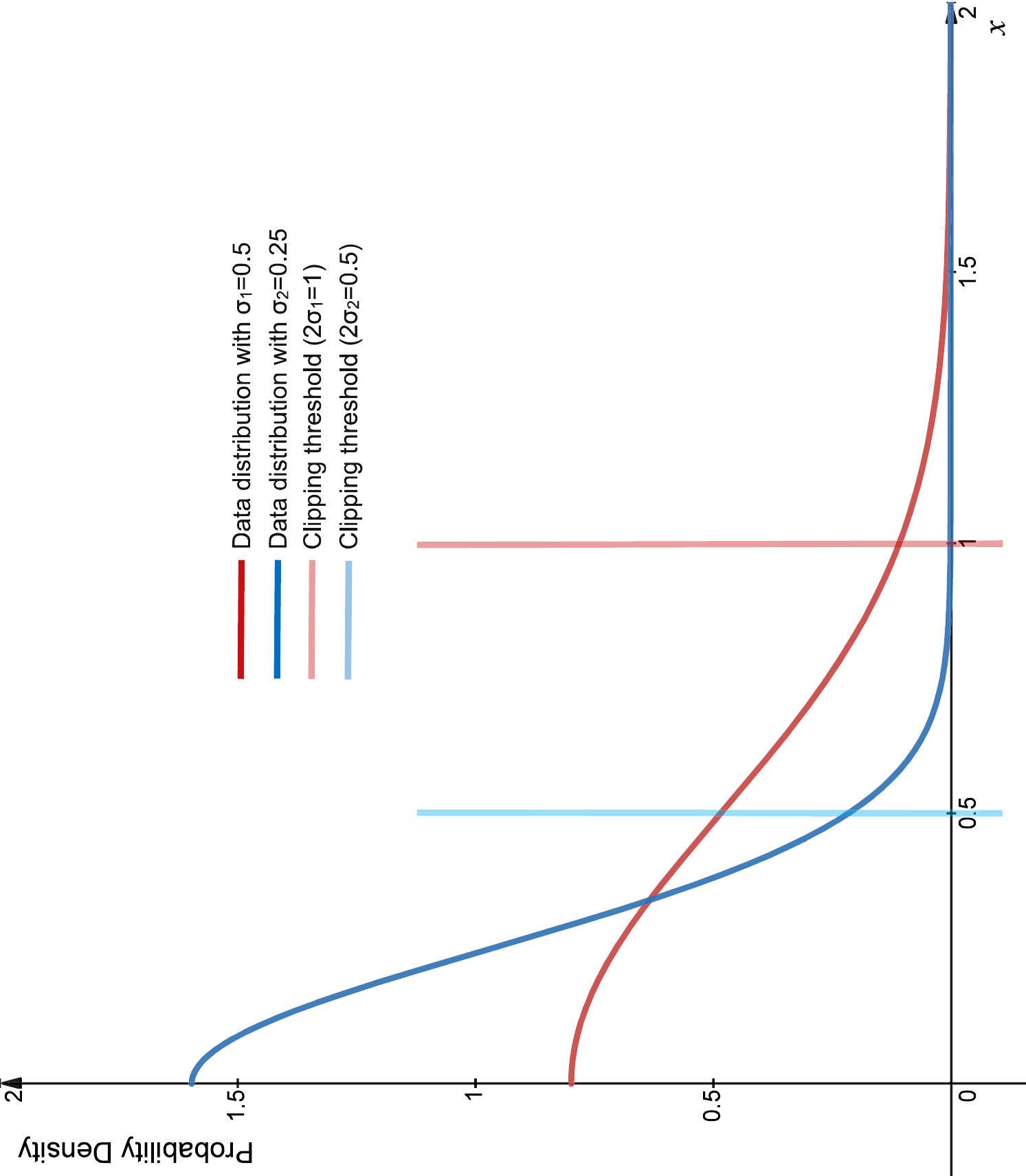}}
\vskip -0.2in
\caption{Clipping thresholds of two set of data with standard deviation of 0.5 and 0.25. The quantizer parameter $\alpha$ is fixed ($\alpha = 2$) for both data distributions.}
\label{fastconverge}
\end{center}
\vskip -0.5in
\end{figure}

\textbf{Faster convergence:}
In the previously-proposed methods, the clipping threshold of weights and activations solely relies on the quantizer parameter $\alpha$. Consequently, if the distribution changes rapidly, it may take a longer time for the clipping threshold to catch up with the new distribution of data. In our proposed quantizer, the clipping threshold relies on both the $\alpha$ and the standard deviation of weights/activations distribution. As a result, even without updating $\alpha$, the clipping threshold can follow the new distribution effortlessly as illustrated in \cref{fastconverge}. % provided in \cref{abs}. %In fact, the standard deviation used in our quantization method gives us a simple understanding of how data is being quantized. In the original PACT quantization method, the clipping threshold, i.e., quantizer parameter $\alpha$, does not convey any information about weights/activations distribution. In PACT, we only know that values larger than the clipping threshold (outliers) are clipped and quantized to the highest quantization level. On the other hand, in our method, we have an estimation of outliers and pruned values by knowing the standard deviation of the weights/activations distribution, although this estimation is not accurate. For instance, let us assume the weights have a Gaussian distribution and $\alpha$ is 2. In this hypothetical example, we know 95.45\% of weights are inside the clipping threshold whereas only 4.55\% are considered as outliers.

%\begin{figure}[t]
%\vskip 0.2in
%\begin{center}
%\centerline{\includegraphics[width=0.7\columnwidth, angle=-90, origin=c]{fastconverge.pdf}}
%\caption{Clipping thresholds of two set of data with standard deviation of 0.5 and 0.25. The quantizer %parameter $\alpha$ is fixed ($\alpha = 2$) for both data distributions.}
%\label{fastconverge1}
%\end{center}
%\vskip -0.4in
%\end{figure}
\vspace{-0.2cm}
\section{Quantization Techniques}\label{Techniques}

In this section, we propose two quantization techniques that can be employed to further improve the performance of our quantization method.

\subsection{Improved Progressive Training}
\begin{figure}[t]
%\vskip -0.1in
\begin{center}
\centerline{\includegraphics[width=0.9\columnwidth]{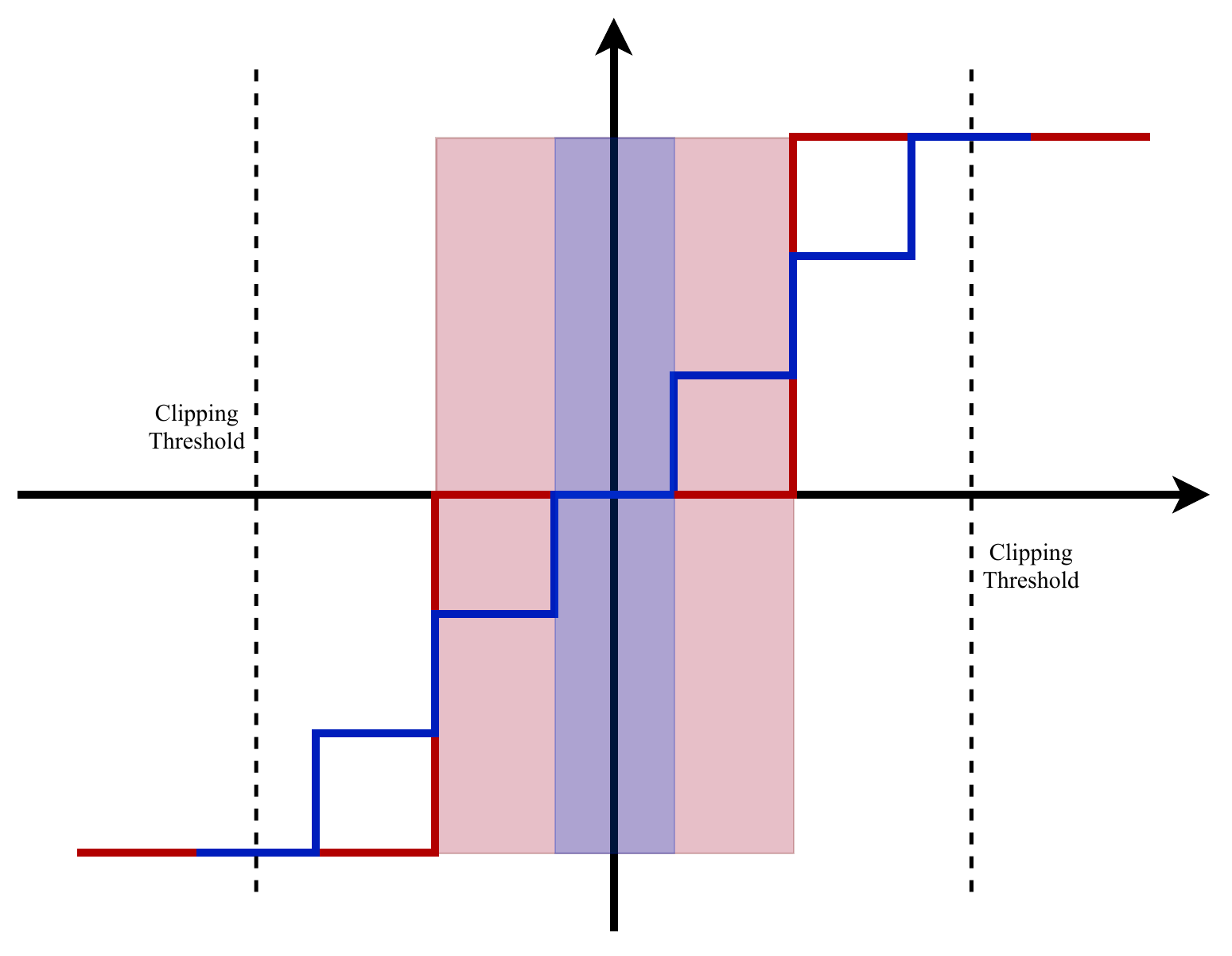}}
\vskip -0.2in
\caption{Quantization levels (intervals) of a 3- and 2-bit quantizers with the same $\alpha$. The blue line shows the intervals of the 3-bit quantizer, whereas the red line shows the intervals for the 2-bit quantizer. The pruning area of the 2-bit quantizer (rectangle with the shade of red) is 3 times wider than the 3-bit quantizer (rectangle with the shade of blue).}
\label{Scaling}
\end{center}
\vskip -0.25in
\vspace{-0.3cm}
\end{figure}

Arguably, any neural network quantized to extremely low bit-widths suffers from accuracy loss due to poor data representation. However, the inaccurate representation of the weights/activations is not the only factor, negatively impacting neural networks. In fact, the gradient vanishing problem as the result of intense pruning during the quantization process is a major issue causing significant performance degradation \cite{s3}. Furthermore, intense pruning reduces the learning capacity of neural networks since a huge portion of the weights are removed (set to zero). %\par
Previous studies tried to address this issue by proposing different progressive training methods \cite{zhou2017incremental, PQTSG}. A recent method, for instance, proposed to quantize higher values of weights first while keeping the lower values in full-precision \cite{zhou2017incremental}. This method allows the gradient to back propagate through the weights with lower values which would have been pruned due to quantization. In another method, it has been shown that a progressive training (e.g., training 2-bit width networks using the trained
parameters from the 3-bit width networks) improves the training capacity and the accuracy performance of the quantized networks. However, by just transferring the learned parameters, we are changing the learned quantization intervals according to \cref{yd}. Consequently, a
huge portion of the network parameters are pruned during the transformation from higher bit-width networks
to lower bit-width networks due to the larger pruning area as illustrated in \cref{Scaling}. By changing the quantization intervals, the
learning capacity of the quantized network will be reduced even more, not only because of the poor
representation of data due to low bit-width quantization, but also as a result of the pruning. \par To address this, we propose to re-scale the quantizer parameter $\alpha$ so that the lower bit-width network starts with the same quantization intervals of the higher bit-width network by applying
\begin{equation}
\alpha_b = \alpha_{b+n} \times \frac{L_b}{L_{b+n}},
\end{equation}
where $\alpha_{b}$ and $L_{b}$ are the quantizer parameter and the discretization level of the quantized networks with $b$ bits, respectively. \par

Our proposed progressive training method offers two advantages: improving the training capacity of the quantized network and preventing the quantizer from pruning the parameters
further. Consequently, our method reduces the impact of gradient vanishing problem. More importantly, this allows us to limit the search space for the quantizer parameter $\alpha$, forcing
the network to find optimal intervals near the ones found in networks with higher bit-widths. This
can be done by using smaller gradient scale values $s$ in \cref{newgrad} which allows the weight decay to prevent the quantizer parameter $\alpha$ from getting larger.

\subsection{Two-Phase Training}

\begin{figure}[t]
%\vskip 0.1in
\vspace{-0.7cm}
\begin{center}
\centerline{\includegraphics[width=0.8\columnwidth]{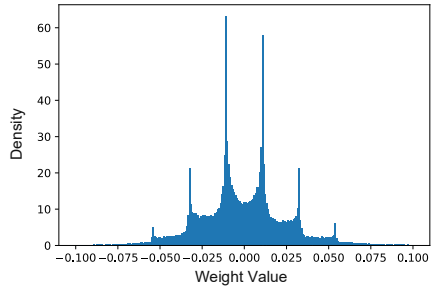}}
\vskip -0.3in
\caption{Distribution of a convolution layer from ResNet-18 model trained with our quantization method using 3 bits. The spikes in the distributions are the transition boundaries of the 3-bit quantizer.}
\label{Weight-dis}
\end{center}
\vskip -0.6in
\end{figure}

It has been shown that a huge portion of the weights resides near the quantization intervals
boundaries (transition boundaries) \cite{qil}. Our experimental result also shows the same patterns across different networks as illustrated in \cref{Weight-dis}. Based on these observations, we suspected that jointly optimizing the network parameters along with the quantizer’s could negatively impact the performance of the network since a small change in quantization intervals can possibly result in a dramatic change in the quantized weights, especially for 2-bit quantized networks where there are fewer discrete levels. In addition, it is more difficult for weights near the transition boundaries to converge to the optimal quantization intervals when the transition boundaries are constantly fluctuating. Furthermore, while we do not see the same pattern for the quantized activations (because of the batch normalization), this fluctuation could still hurt the quantization of the activation but with less
severity. Furthermore, the gradients of the quantized networks with learnable parameter are affected
by that parameter when passing through the layers. Looking at \cref{lossx}, it is evident that both $\alpha$ and $\sigma$ are controlling the gradients passing to the previous layers. Since $\alpha$ and $\sigma$ are both getting updated
continuously during the training, we suspected that this might introduce some noise to the gradients. To investigate this hypothesis, we retrained our quantized networks keeping $\alpha$ and $\sigma$ frozen
to their optimal value from the first training. Interestingly, our empirical results show constant improvement across different networks and datasets, which highlights the importance of our proposed
two-phase training method when training parameterized quantized networks.

%\subsection{Modified L2-regularization}
%The functionality of our proposed method heavily relies on using the L2-regularization method on the weight parameters. Without using L2-regularization, the weight distributions become scattered and our method fails to efficiently perform quantization. On the other hand, by using L2-regularization, a large amount of weights are forced to move towards the center of the distribution and consequently, these weights are trapped in the pruning area of the quantizer. Applying L2-regularization on these values does not impact the model performance since these weights are already pruned and do not contribute during forward propagation. In fact, in this scenario, L2-regularization negatively forces the already pruned weights to further move towards zero and eliminates the possibility of exiting the pruning area for these weights. To address this issue, we remove the weight decay (L2-regularization) on weights that reside in the pruning area of the quantizer. According to our experimental results, this simple modification on L2-regularization, significantly improves the performance accuracy of the models quantized to 2 and 3 bits.     

% Note use of \abovespace and \belowspace to get reasonable spacing
% above and below tabular lines.

\begin{table*}[h]
%\vskip -0.15in
\caption{Quantization accuracy performance on CIFAR10 dataset with ResNet-20 model (FP accuracy: 91.74\%). Methods included in this table are LQ-Net \cite{lqnet}, DSQ \cite{dsq} and PACT \cite{pact}}
\label{resnet20}
%\vskip 0.15in
\begin{center}
\begin{small}
\begin{tabular}{lllcccc}
\toprule
\multicolumn{2}{c}{Quantization method} &
  \multirow{2}{*}{Training technique} &
  \multicolumn{4}{c}{Accuracy @ precision (A and W)} \\ 
  \cmidrule{1-2} \cmidrule{4-7} 
Activations & Weights &                       & 5     & 4     & 3     & 2     \\
\midrule
DSQ      & DSQ  & NA                               & --    & --    & --  & 90.11  \\
LQ-Net      & LQ-Net  & NA                   & --    & --    & 91.6  & 90.2  \\
\midrule
PACT        & DoReFa  & From scatch          & 91.7  & 91.3  & 91.1  & 89.7  \\
\midrule
\multirow{3}{*}{Ours} &
  DoReFa &
  From scatch &
  %\multirow{3}{*}{91.74} &
  91.75 &
  91.63 &
  91.25 &
  88.77 \\
            & DoReFa  & Progressive               & 91.90 & 91.81 & 91.68 & 89.74 \\
            & DoReFa  & Two-phase training        & 92.03 & 92.00 & 91.65 & 90.32 \\
\midrule
Ours &
  Ours &
  \begin{tabular}[c]{@{}l@{}}Progressive with re-scaling\\ and two-phase training\end{tabular} &
   
  \textbf{92.27} &
  \textbf{92.28} &
  \textbf{92.23} &
  \textbf{90.77} \\
  \bottomrule
\end{tabular}
%\end{sc}
\end{small}
\end{center}
\vskip -0.2in
\end{table*}

\vspace{-0.3cm}
\section{Experiments}
To validate our proposed quantization method, we conduct several experiments on CIFAR10 \cite{krizhevsky2009learning} and ImageNet \cite{ILSVRC15} datasets on various architecture. Furthermore, we present several ablation studies to evaluate the impact of the hyper-parameters and the proposed quantization techniques on the performance of the quantized networks. In all experiments, we use the network's weight decay value for the quantizer's weight decay. Furthermore, we use the same data augmentation proposed in \cite{resnet} for both CIFAR10 and ImagNet datasets.

\subsection{ResNet-20 on CIFAR10}

We demonstrate the effectiveness of our proposed method by quantizing ResNet-20 model \cite{resnet} on CIFAR10 dataset. To have a fair comparison with PACT, we conduct two different experiments based on the quantization method applied on weights. In the first experiment, we quantize ResNet-20 model progressively with re-scaling of the clipping threshold and applying the proposed two-phase training technique. In this experiment, both weights and activations are quantized using our proposed quantization scheme. The gradient scale value ($s$) is set to \{1, 1, 0.1, 0.01\} for the \{5, 4, 3, 2\}-bit quantized ResNet-20, respectively. It is worth mentioning that the gradient scale values were adjusted for each bit-width after performing hyper-parameter search (see \cref{hyperparameter}). \par   
In the second experiment, we employ the same weight quantization method used in PACT.  We also quantize ResNet-20 from scratch, and provide the accuracy results separately in \cref{resnet20}. Finally, we employ the progressive method \textbf{without re-scaling} and our proposed two-phase training technique step by step to demonstrate their impact on the performance of the quantized network. Similar to PACT, we do not quantize the first layer and the last layer of ResNet-20. In this experiment, the gradient scale used to quantize activations is set to 1 regardless the bit-widths.      

From \cref{resnet20}, we observe that our quantization method achieves a higher accuracy compared to PACT even when only the activations are quantized with our method. Furthermore, we achieve a higher accuracy when our two-phase training technique is applied over the progressive training method, showing its effectiveness in reducing the gradient noise. In addition, we achieve the best performance when both weights and activations are quantized using our method across all the quantization bit-widths.

\subsection{SmallVGG on CIFAR10}
We also evaluate our method on CIFAR10 dataset by quantizing the SmallVGG network \cite{rqst} under three different setups:\\
\textbf{Setup-1} We use our quantization method to quantize both weights and activations of SmallVGG network with 2 bits. In this setup, all layers except the first convolution layer and the fully-connected layer are quantized. The network's parameters are initialized with the full-precision parameters. We train the network for 300 epochs using gradient scale of 0.001. As shown in \cref{smallvgg}, our method outperforms the state-of-the-art and even achieves a higher accuracy performance compared to the full-precision model.\\
\textbf{Setup-2} In this setup, we quantize activations to 2 bits (similar to the first experiment), but we binarize weights (i.e., quantize with 1 bit) using the sign function as described in \cite{courbariaux2016binarized}. Interestingly, our method still manages to outperform SOTA and full-precision while using binarized weights and 2-bit activations during inference.\\  
\textbf{Setup-3} We repeat the same experiment described in setup-1 to quantize all layers of SmallVGG network. It should be noted that only the weights of the first layer are quantized while its inputs are kept in full-precision. Similar to the first experiment, our method outperformed SOTA and full-precision models. The results from these three experiments show quantization can have a regularization effect as long as the capacity of the quantized network is not decreased due to quantization. Furthermore, we can conclude that SmalllVGG network is over-parameterized for small datasets like CIFAR10.

\begin{table}[th]
\vskip -0.15in
\caption{Comparison with SOTA methods on CIFAR10 dataset using SmallVGG network (FP accuracy: 93.66\%). Methods included in this table are LQ-Net \cite{lqnet}, HWGQ \cite{HWGQ}, LLSQ \cite{llsq} and RQST \cite{rqst}.}
\label{smallvgg}
%\vskip 0.15in
\begin{center}
\begin{small}
%\begin{sc}

\begin{tabular}{lccc}
\toprule
\multirow{2}{*}{Method} & \multirow{2}{*}{All layers} & \multicolumn{2}{r}{Accuracy @ precision (A/W)} \\
\cmidrule{3-4}
       &      & 2/1            & 2/2            \\
\midrule
LQ-Net & No    & 93.40           & 93.50           \\
HWGQ   & No   & 92.51          & NA             \\
LLSQ   & No   & NA             & 93.31          \\
Ours   & No   & \textbf{93.88} & \textbf{94.36} \\
\midrule
RQST   & Yes  & NA             & 90.92          \\
LLSQ   & Yes  & NA             & 93.12          \\
Ours   & Yes  & NA             & \textbf{93.90} \\
\bottomrule
\end{tabular}

%\end{sc}
\end{small}
\end{center}
\vskip -0.3in
\end{table}

%\vspace{-0.4cm}
\subsection{AlexNet on ImageNet}
We test our quantization method on ImageNet dataset using modified AlexNet \cite{krizhevsky2012imagenet} where a batch normalization layer was added after convolution and fully-connected layers except for the last layer. We use the progressive training method with re-scaling when training the 3-bit and the 2-bit networks. The quantized network is optimized with cosine annealing scheduler with the initial learning rate of 0.001 for 70 epochs. Following the common practice from previous methods, all layers  are quantized except for the first layer and the last layer. As shown in \cref{alexnet}, our proposed method outperforms the previously-proposed quantization methods. 

\begin{table}[b]
\vskip -0.3in
\caption{Comparion with the existing methods on AlexNet (FP accuracy: 61.8\%). Methods included in this table are QIL \cite{qil}, LQ-Net \cite{lqnet} and TSQ \cite{tsq}.}%, SYQ \cite{faraone2018syq}, PACT \cite{pact}, LLSQ \cite{llsq}, BalancedQ \cite{zhou2017balanced}, PQTSG \cite{PQTSG} and WEQ \cite{park2017weighted}.}
\label{alexnet}
%\vskip 0.15in
\begin{center}
\begin{small}
%\begin{sc}

\begin{tabular}{lccc}
\toprule
\multirow{2}{*}{Method} & \multicolumn{3}{c}{Top-1 accuracy @ precision (A and W)} \\
\cmidrule{2-4}
                        & 4              & 3              & 2            \\
                        \midrule
Ours                    & \textbf{62.5}           & \textbf{62.2}           & \textbf{59.2}         \\
QIL                     & 62             & 61.3           & 58.1         \\
LQ-Net                  & --             & --             & 57.4          \\
TSQ                     &  --            & --             & 58           \\
%SYQ                     &  --            &  --            & 55.8         \\
%PACT                    & 57.2           & 55.6           & 55.0         \\
%LLSQ                    & 56.57          & 55.36          &  --          \\
%BalancedQ               &   --           & --             & 55.7         \\
%PQTSG                   & 58.1           & --             & 52.5         \\
%WEQ                     & 55.9           & 54.9           & 50.6        \\
\bottomrule
\end{tabular}

%\end{sc}
\end{small}
\end{center}
\vskip -0.15in
\end{table}

%\subsubsection{2-bit Quantized AlexNet}

When training the 2-bit AlexNet with the gradient scale of 1 (i.e., $s = 1$), we observed that the training loss starts to increase after several epochs and the network converges to a poor local minima due to extreme pruning, even with re-scaling of the clipping threshold. To address this issue, we changed $s$ to 0.01 to force the 2-bit AlexNet use the intervals found from the 3-bit network with less weight pruning. Furthermore, we trained the 2-bit AlexNet with the intervals obtained from the 4-bit network with $s = 0.01$. Finally, we used our double training method for the 2-bit AlexNet which increased the accuracy by 0.2\%. As shown in \cref{alexnet-ab}, the 2-bit AlexNet with 3-bit intervals has the best performance accuracy while the one trained with 4-bit intervals also outperforms the QIL method.

\begin{table}[ht]
\vskip -0.15in
\caption{Top-1 accuracy of 2-bit AlexNet under different training setup.}
\label{alexnet-ab}
%\vskip 0.15in
\begin{center}
\begin{small}
%\begin{sc}

\begin{tabular}{lcc}
\toprule
\multirow{2}{*}{Setup}                                                           & \multirow{2}{*}{Gradient scale} & \multirow{2}{*}{Top-1 Acc. (\%)} \\
                                                                                 &                                 &                                  \\
                                                                                 \midrule
No re-scaling                                                                    & 1                               & 54.4                             \\
\midrule
\multirow{2}{*}{\begin{tabular}[c]{@{}l@{}}Re-scaled \\ from 4-bit\end{tabular}} & \multirow{2}{*}{0.01}           & \multirow{2}{*}{58.76}           \\

                                                                                 &                                 &                                  \\
                                                                                 \midrule
\multirow{2}{*}{\begin{tabular}[c]{@{}l@{}}Re-scaled \\ from 3-bit\end{tabular}} & \multirow{2}{*}{0.01}           & \multirow{2}{*}{59.01}           \\
                                                                                 &                                 &                                  \\
                                                                                 \midrule
\multirow{3}{*}{\begin{tabular}[c]{@{}l@{}}Re-scaled \\ from 3-bit + \\ two-phase training\end{tabular}} & \multirow{3}{*}{0.01} & \multirow{3}{*}{59.24} \\
                                                                                 &                                 &                                  \\
                                                                                 &                                 &           \\
                                                                                 \bottomrule
\end{tabular}

%\end{sc}
\end{small}
\end{center}
\vskip -0.3in
\end{table}

%\subsection{The impact of L2-regularization}
%In order to evaluate the impact of our improved regularization technique, we retrain a 2-bit quantized ResNet-20 on CIFAR10 dataset with the original L2-Regularization. As shown in Table~\ref{}, the model trained with our modified regularization technique achieved a significantly better accuracy performance compared to the one trained with the original L2-Regularization. In addition, our modified regularization technique resulted in lower pruning ratio, proving that our technique helps preventing the weight parameters from trapping inside the pruning area of the quantizers. Consequently, our regularization technique provides higher learning capacity when models are quantized to extremely low bits e.g., 2-bit.
%\vspace{-0.2cm}
\subsection{ResNet on ImageNet} \label{non-uniform}
We evaluate our proposed base-2 logarithmic quantization method described in \cref{shiftnet} on ImageNet dataset using ResNet-18 and ResNet-50 models \cite{resnet}. The under-test models are specifically chosen to fairly compare the accuracy performance of our method with SOTA  shift-add methods. We employ previously described two-phase training technique. Both models are initialized with the pre-trained parameters from the Pytorch model zoo \cite{NEURIPS2019_9015}. We train the models for 70 epochs with cosine learning scheduler. The results summarized in \cref{shiftnet-res} show that our method outperforms the SOTA results for both models. In addition to having higher accuracy performance, our method has several advantages over the Sign-Sparse-Shift \cite{s3} method. First, the full-precision parameters of the pre-trained models can be used to initialize the quantized models. Consequently, the quantized models can converge to an acceptable accuracy much faster than the ones initialized randomly. In fact, the Sign-Sparse-Shift method requires 200 epochs to reach the results provided in \cref{shiftnet-res}, whereas our method achieves a higher accuracy after only 20 epochs (10$\times$ faster convergence). Furthermore, the Sign-Sparse-Shift method requires 4 times the number of parameters of its full-precision model during the training phase, which significantly increases the training hardware memory footprint and utilization. Contrary, our method only adds a small number of parameters and registers i.e., $\alpha$ and the standard deviation of the weights.

\begin{table}[ht]
\vskip -0.1in
\caption{Comparison between existing shift-add networks on ImageNet dataset. Methods included in this table are DeepShift \cite{elhoushi2021deepshift}, INQ \cite{zhou2017incremental} and Sign-Sparse-Shift ($S^3$) \cite{s3}.}
\label{shiftnet-res}
\vskip -0.15in
\begin{center}
\begin{small}
%\begin{sc}

\begin{tabular}{llcc}
\toprule
Model & Method & Width & \begin{tabular}[c]{@{}c@{}}Top-1/Top-5 \\ Acc. (\%)\end{tabular} \\
\midrule
\multirow{8}{*}{ResNet-18} & FP                   & 32 & 69.76/89.08          \\
                           & DeepShift            & 5  & 69.56/89.17          \\
                           & INQ                  & 3  & 68.08/88.36          \\
                           & $S^3$                & 3  & 69.82/89.23          \\
                           & Ours                 & 3  & \textbf{70.04/89.14} \\    
                           & INQ                  & 4  & 68.89/89.01          \\
                           & $S^3$                & 4  & 70.47/89.93          \\
                           & Ours                 & 4  & \textbf{70.70/89.62} \\
\midrule                           
\multirow{5}{*}{ResNet-50} & FP                   & 32 & 76.13/92.86          \\
                           & INQ                  & 5  & 74.81/92.45          \\
                           & DeepShift            & 5  & 76.33/93.05          \\
                           & $S^3$                & 3  & 75.75/92.80          \\
                           & Ours                 & 3  & \textbf{76.41/93.01} \\
\bottomrule                           
\end{tabular}

%\end{sc}
\end{small}
\end{center}
\vskip -0.2in
\end{table}

\subsection{Pruning Ratio and Accuracy Trade-off} \label{Pruningratio}
\begin{figure}[t]
\vskip 0.1in
\begin{center}
\centerline{\includegraphics[width=\columnwidth]{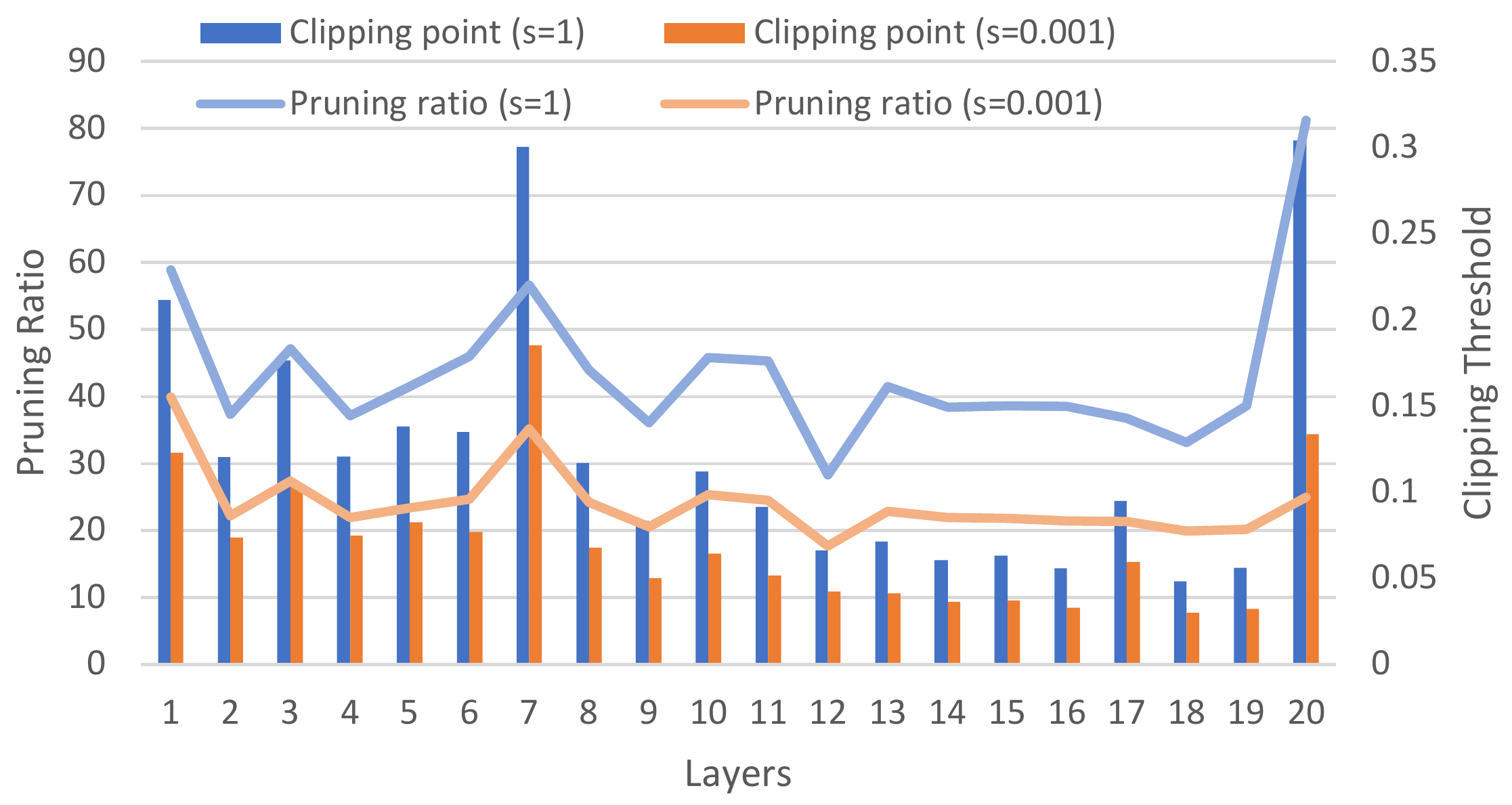}}
\vskip -0.2in
\caption{The clipping thresholds and the pruning rates of the quantized ResNet-18 weights}
\label{pruneVSth}
\end{center}
\vskip -0.4in
\end{figure}

As mentioned previously, pruning is an inherit outcome of quantization. A well-pruned network can significantly reduce the size of the memory. However, too much pruning has a negative impact on the model learning capacity. In our quantization method, we can control the pruning ratio indirectly by adjusting the gradient scale ($s$ in \cref{newgrad}). We evaluate this property by training the same 3-bit quantized ResNet-18 model from previous experiment (\cref{non-uniform}) on ImageNet dataset with various gradient scale factors. Each model was trained for 20 epochs. The results provided in \cref{PRUNEvsACC} show that we can achieve slightly better accuracy performances at the cost of loosing a significant amount of pruning ratio. More specifically, we can achieve 1.12\% higher accuracy if we accept 18.64\% less pruning. It is up to the users to decide whether the benefits of networks with higher pruning ratios outweigh their slight drop in accuracy. It should be noted that this trade-off can only be considered in a situation where several gradient scale values result in a good convergence and accuracy performance. Furthermore, the clipping thresholds and the pruning rates of each layers for the quantized networks with the gradient scale values of 1 and 0.001 are shown in \cref{pruneVSth} to better demonstrate how the gradient scale affects the pruning ratio. As expected, the quantized network with the smaller gradient scale value has smaller clipping thresholds. Furthermore, higher clipping thresholds result in grater pruning rates.

\begin{table}[t]
%\vskip -0.1in
\caption{Top-1 accuracy and pruning ratio for various gradient scale factors.}
\label{PRUNEvsACC}
%\vskip 0.15in
\begin{center}
\begin{small}
%\begin{sc}

\begin{tabular}{lcccc}
\toprule
Gradient scale     & 1     & 0.1   & 0.01  & 0.001 \\
\midrule
Top-1 Acc. (\%)    & 68.92 & 69.59 & 69.70 & 70.04 \\
Pruning ratio (\%) &  40.21     &   29.74    &   24.97    &   21.57    \\
\bottomrule
\end{tabular}

%\end{sc}
\end{small}
\end{center}
\vskip -0.1in
\end{table}

\vspace{-0.2cm}
\subsection{Progressive Training and Re-scaling} \label{hyperparameter}

%We evaluate the effectivness of our proposed progressive training, the re-scaling of the quantizer clipping threshold of the weight parameters, by training ResNet-20 and AlexNet models on CIFAR10 and ImageNet datasets, respectively. In this experiment, the models are trained with and without re-scaling. From Table~\ref{}, we observe that the models trained with re-scaling, significantly outperform the one trained without re-scaling, specially, when models are quantized to 2 bits. The gap between the accuracy performance increases even more for the AlexNet model which is trained on the more challenging and bigger dataset, i.e., ImageNet. The results of this experiment support our hypothesis that using re-scaling can prevent the gradient vanishing problem in quantized network if combined with progressive training.

We evaluate the effectiveness of our proposed progressive training, i.e., the re-scaling of the clipping threshold for the weights quantizer, by training ResNet-20 on CIFAR10 dataset. In this experiment, ResNet-20 model is quantized to 2 and 3 bits. The 2-bit quantized model is initialized and re-scaled from the 3-bit and 4-bit models, and the 3-bit model is initialized and re-scaled using the 4-bit model. From \cref{rescalingtable}, we observe that the 2-bit model yields a higher accuracy when the clipping threshold is re-scaled from the 3-bit model. Interestingly, both the 2-bit and the 3-bit models produce comparable accuracy even when the quantizer parameter ($\alpha$) is not updated (the gradient scale is 0). In other words, we can use the same quantization intervals, found from the higher-bit networks, in the lower-bit networks and still achieve acceptable performance. This observation shows the importance and effectiveness of our proposed re-scaling. Furthermore, from the columns where the gradient scale is 0.1 and 0.01, we observe that updating $\alpha$ with a smaller gradient scale does not necessarily achieve a better performance all the time.

%trained with re-scaling, significantly outperform the one trained without re-scaling, specially, when models are quantized to 2 bits. The gap between the accuracy performance increases even more for the AlexNet model which is trained on the more challenging and bigger dataset, i.e., ImageNet. The results of this experiment support our hypothesis that using re-scaling can prevent the gradient vanishing problem in quantized network if combined with progressive training.

\begin{table}[t]
\vskip -0.1in
\caption{Accuracy performance of ResNet-20 on CIFAR10, quantized using 2 and 3 bits with different clipping threshold initialization and gradient scale values.}
\label{rescalingtable}
\vskip -0.1in
\begin{center}
\begin{small}
%\begin{sc}

\begin{tabular}{cccccc}
\toprule
\multirow{3}{*}{Bit-width} &
  \multirow{3}{*}{\begin{tabular}[c]{@{}c@{}}Clipping \\ Threshold\\ Initialization\end{tabular}} &
  \multicolumn{4}{c}{\multirow{2}{*}{Gradient scale}} \\
  &        & \multicolumn{4}{c}{}          \\
  \cmidrule{3-6}
  &        & 1     & 0.1   & 0.01  & 0     \\
 \midrule
2 & From 3 & 88.06 & 90.34 & 90.69 & 90.12 \\
2 & From 4 & 87.55 & 90.29 & 90.11 & 89.57 \\
3 & From 4 & 92.14 & 92.23 & 92.07 & 92.10 \\
\bottomrule
\end{tabular}

%\end{sc}
\end{small}
\end{center}
\vskip -0.3in
\end{table}

\vspace{-0.2cm}
\section{Conclusion}

We proposed a new quantization method that takes advantage of the knowledge of weights and activations distribution during the quantization process. Using the standard deviation of weights and activation, our proposed method outperforms the previous works on various image classification tasks. Furthermore, we proposed two training techniques to further improve the performance of our quantization scheme. We introduced a two-phase training technique to address the gradient noise and fluctuation of the quantizer's transition boundaries caused by the joint optimization of the network's and the quantizer's parameters. In addition, we presented a novel re-scaling technique to improve the training of quantized networks by reducing the impact of gradient vanishing problem. Finally, we proposed a novel non-uniform quantization framework (base-2 logarithmic quantization) where weights are quantized to power-of-two discrete intervals to replace expensive multipliers with simple shift-add operations. The proposed base-2 logarithmic quantization method converges 10$\times$ faster that the SOTA method and outperforms existing work. Our proposed quantization method provides the flexibility to trade-off between accuracy and network size by controlling the pruning ratio. In future work, we plan to investigate the compatibility of our method with transformers.

% \section*{Acknowledgements}

% \textbf{Do not} include acknowledgements in the initial version of
% the paper submitted for blind review.

% If a paper is accepted, the final camera-ready version can (and
% probably should) include acknowledgements. In this case, please
% place such acknowledgements in an unnumbered section at the
% end of the paper. Typically, this will include thanks to reviewers
% who gave useful comments, to colleagues who contributed to the ideas,
% and to funding agencies and corporate sponsors that provided financial


\bibliography{example_paper}
\bibliographystyle{icml2022}

%%%%%%%%%%%%%%%%%%%%%%%%%%%%%%%%%%%%%%%%%%%%%%%%%%%%%%%%%%%%%%%%%%%%%%%%%%%%%%%
%%%%%%%%%%%%%%%%%%%%%%%%%%%%%%%%%%%%%%%%%%%%%%%%%%%%%%%%%%%%%%%%%%%%%%%%%%%%%%%
% APPENDIX
%%%%%%%%%%%%%%%%%%%%%%%%%%%%%%%%%%%%%%%%%%%%%%%%%%%%%%%%%%%%%%%%%%%%%%%%%%%%%%%
%%%%%%%%%%%%%%%%%%%%%%%%%%%%%%%%%%%%%%%%%%%%%%%%%%%%%%%%%%%%%%%%%%%%%%%%%%%%%%%
%\newpage
%\appendix
%\onecolumn
%\section{Comparison Between Ours and PACT Clipping Functions} \label{appendix}

%The standard deviation used in our quantization method gives us a simple understanding of how data is being quantized. In the original PACT quantization method, the clipping threshold, i.e., quantizer parameter $\alpha$, does not convey any information about weights/activations distribution. In PACT, we only know that values larger than the clipping threshold (outliers) are clipped and quantized to the highest quantization level. On the other hand, in our method, we have an estimation of outliers and pruned values by knowing the standard deviation of the weights/activations distribution, although this estimation is not accurate. For instance, let us assume the weights have a Gaussian distribution and $\alpha$ is 2. In this hypothetical example, we know 95.45\% of weights are inside the clipping threshold whereas only 4.55\% are considered as outliers.
%%%%%%%%%%%%%%%%%%%%%%%%%%%%%%%%%%%%%%%%%%%%%%%%%%%%%%%%%%%%%%%%%%%%%%%%%%%%%%%
%%%%%%%%%%%%%%%%%%%%%%%%%%%%%%%%%%%%%%%%%%%%%%%%%%%%%%%%%%%%%%%%%%%%%%%%%%%%%%%

\end{document}